# Monitoring Time-Varying Changes of Historic Structures Through Photogrammetry-Driven Digital Twinning

Xiangxiong Kong

Dept. of Civil and Geomatics Engineering, California State University, Fresno, 2320 E. San Ramon Ave., Fresno, CA 93740, United States - xkong@csufresno.edu

**Keywords:** Heritage Preservation, Photogrammetry, Point Cloud Processing, Unmanned Aerial Vehicle, Digital Twin, Cracks.

**Abstract**

Historic structures are important for our society but could be prone to structural deterioration due to long service durations and natural impacts. Monitoring the deterioration of historic structures becomes essential for stakeholders to take appropriate interventions. Existing work in the literature primarily focuses on assessing the structural damage at a given moment instead of evaluating the development of deterioration over time. To address this gap, we proposed a novel five-component digital twin framework to monitor time-varying changes in historic structures. A testbed of a casemate in Fort Soledad on the island of Guam was selected to validate our framework. Using this testbed, key implementation steps in our digital twin framework were performed. The findings from this study confirm that our digital twin framework can effectively monitor deterioration over time, which is an urgent need in the cultural heritage preservation community.

## 1. Introduction

Historic structures are critical for our lives and society, bridging our past, present, and future, and fostering a deep sense of identity. The preservation of historic structures, therefore, is essential for government agencies, local communities, and other stakeholders to maintain this tangible connection to our society.

Recent advances in photogrammetry technology integrated with Unmanned Aerial Vehicles (UAVs) show great promise in detecting structural deterioration of historic structures. For instance, Galantucci et al. (2019) developed a photogrammetry 3D model of an Italian historic building, producing orthophotos of its limestone façade to identify structural deterioration from cavities. Forlin et al. (2018) utilized photogrammetry on archaeological sites in Cyprus and Spain, extracting orthophotos from 3D models to detect earthquake-induced crack patterns on the buildings' elevation walls. Ulvi (2022) utilized UAVs and photogrammetry to generate 3D dense point clouds of a Turkish archaeological site across four excavation phases, from which elevation profiles of specific cut sections were derived and comparatively analyzed.

While these investigations have yielded successful outcomes, these methods primarily focus on evaluating the status of the structures at a given moment instead of assessing the developments of deteriorations over time (see Figure 1). Although an engineering team can conduct structural inspection across multiple visits using the same method mentioned above, a critical challenge lies in understanding the evolution of structural deterioration between different inspections (e.g., crack propagation, ongoing settlement, continuous decay). This highlights the importance of investigating new research frameworks to monitor time-varying structural deteriorations.

This paper proposes a photogrammetry-driven digital twin framework and illustrates its feasibility for monitoring time-varying deterioration in historic structures. Our approach is built upon two existing knowledge domains: 1) photogrammetry techniques in historic preservation; and 2) proven capabilities of digital twin applications from civil and construction engineering disciplines. By integrating these two domains, we hope to broaden digital twin applications in heritage preservation.

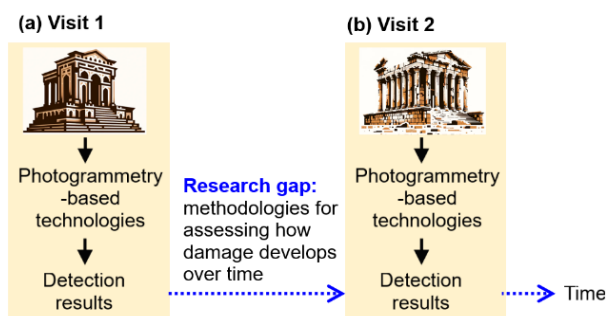

Figure 1. Research gap to be addressed in this study

The rest of the paper is structured as follows: Section 2 discusses the proposed digital twin framework and explains its key methodological principles; Section 3 describes the testbed selection; Sections 4 to 6 illustrate validations of the key feedback flows in the proposed framework including virtual entity reconstruction in Phase 1, deterioration monitoring in Phase 2, and decision-making in Phase 3, Section 7 differentiates our digital twin with those proposed by others, Section 8 concludes the study, and Section 9 summarizes the future work.

## 2. Digital Twinning

The scientific discovery of the digital twin has exponentially grown recently, spanning aerospace, manufacturing, biomedical, construction engineering, environmental science, and healthcare (Jiang et al., 2021). Grieves and Vickers (2016) describe a digital twin as a comprehensive collection of virtual information that represents a physical manufactured product in detail. Later on, Jiang et al. (2021) reviewed digital twin applications in civil engineering based on 134 studies and proposed a definition of digital twin encompassing five key components: 1) the real-world physical entity; 2) the virtual entity that mirrors the physical entity; 3) the connections that enable data transfer; 4) the data repository; and 5) the service that enhances specific objectives for the system, as shown in Figure 2a. All the connections (indicated as black arrows in the figure) are bidirectional flows except the optional feedback flow from data to the physical entity, as suggested by the authors.

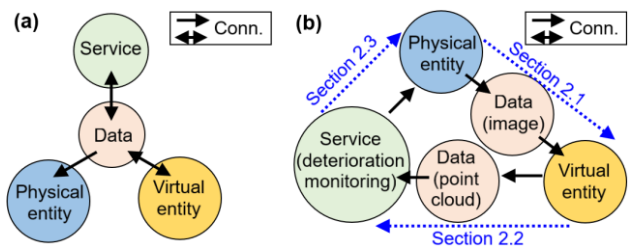

Figure 2. Comparison of digital twin frameworks: (a) proposed by Jiang et al., 2021; and (b) proposed in this study.

Due to the limited digital twin frameworks reported in the current literature on cultural heritage, a new framework is proposed in this study as shown in Figure 2b. Our framework is built upon the digital twin concept defined by Jiang et al. (2021), mainly because our research goal, monitoring structural deteriorations of historic structures, is closely related to the digital twin applications in construction engineering reviewed by Jiang et al. (2021). To better suit the photogrammetry technological workflow, we reconfigured some key components in our framework: 1) the data component between service and physical entity is removed in our definition, and 2) all connections are designed as a unidirectional loop. The methodological principles for all the feedback flows in our framework will be explained in the rest of this section.

**2.1 Feedback Flow: from the Physical Entity to its Virtual Entity**

In this study, we define the physical entity as any historic structure of interest such as buildings, bridges, arenas, towers, or fortifications; and the virtual entity as a 3D dense point cloud model of the selected physical entity. To reconstruct the point cloud, many existing methodologies can fit this purpose such as LiDAR/laser scanners-based (Castagnetti et al., 2012) or Structure-from-Motion Multi-View-Stereo (SfM-MVS)-based (Brandolini and Patrucco, 2019). This study applies a photogrammetry workflow based on SfM-MVS. The data to construct the point cloud, in this regard, is a large volume of high-resolution images collected by UAVs.

**2.2 Feedback Flow: from the Virtual Entity to Service**

Once the 3D dense point cloud is established in Section 2.1, the point cloud is then transferred from the virtual entity component to the service component. In this study, we define the service component as evaluating the time-varying deterioration of a historic structure. To fulfill this need, cloud-to-cloud (C2C) distances are computed to uncover differential features of the physical entity over time. Computing C2C distances is commonly used in remote sensing (Kong, 2021; Alazmi and Seo, 2023). Lastly, identifying changes over time also requires comparing two point cloud datasets at different inspections. This can be achieved by collecting UAV images of the structure twice for constructing two point clouds.

**2.3 Feedback Flow: from Service to Physical Entity**

Based on the findings from the deterioration monitoring explained in Section 2.2, stakeholders of the historic structure can make informed decisions on possible interventions. For example, if the monitoring results indicate that the structure is safe with no significant deterioration, then no major actions are required. However, if the structure has evidence of substantial deteriorations (e.g., critical crack propagation, severe deformation, or extensive corrosion) that would weaken the structural integrity, then this information will be reported to the stakeholders for taking possible actions such as retrofit, rehabilitation, or replacement of the affected parts of the historic structure.

**2.4 Feedback Loop: Implementation Over Time**

While Sections 2.1 to 2.3 explain a typical digital twin cycle, it is important to notice that the physical entity in the digital twin framework is subjected to time-varying physical states (i.e., ongoing development of structural deterioration). To continuously monitor the time-varying deteriorations over the long term, we can iteratively execute the digital twin loop, repeating the methodological flows described in Sections 2.1 to 2.3. This involves performing further field visits for image collections; building updated virtual models; monitoring additional potential deteriorations, and making informed decisions based on the new data.

**2.5 Our Framework vs. 4D Modeling**

In heritage preservation, 4D modeling refers to combining a physical entity's three spatial dimensions with its temporal dimension for mapping changes over time (Kersten et al., 2014). Although both our framework and 4D modeling adopt temporal approaches, they are different in two notable ways: 1) 4D modeling typically examines changes in the physical entity over extended periods (e.g., decades or centuries), often to hypothesize about past historical developments (Rodríguez-Gonzálvez, et al., 2017). In contrast, our framework focuses on the current state of the historic structure over shorter time intervals, aiming to monitor structural deteriorations in the future. 2) 4D Modeling generally documents the structure at a given time or environmental condition without any feedback mechanism from the virtual entity to the physical entity (Hassan and Fritsch, 2019, Kersten et al., 2014). On the other hand, our framework enables the feedback loop, where the information generated from the virtual entity leads to human interventions applied to the physical entity.

### 3. Testbed

To evaluate our methodology, we selected a historic casemate in Fort Nuestra Señora de la Soledad (Fort of Our Lady of Solitude, hereafter referred to as Fort Soledad) in Umatac, Guam. Guam is the largest island of the Marianas Chain in the Western Pacific, known for its ancient history and rich cultural heritage. Fort Soledad is one of the four fortifications built by the Spanish from 1680 to 1810. Figure 3 shows Fort Soledad and the casemate under different views.

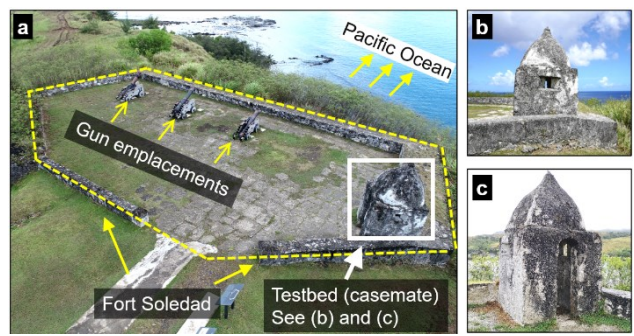

Figure 3. (a) Fort Soledad; (b) and (c) are different views of the casemate, the testbed in this study.

## 4. Validation Phase 1: Virtual Entity Reconstruction

This section validates the feedback flow that links the physical entity and the virtual entity. We first discuss the UAV image collection strategies in Section 4.1; then we explain the procedure for creating the virtual entity of the casemate in Section 4.2.

### 4.1 UAV Operation and Image Collection

To collect images of the testbed, DJI Phantom 4 Pro+ V2.0 (DJI Phantom 4 hereafter) was adopted during the field visit. A total of 395 UAV images were collected of Fort Soledad under different camera positions. Two flight patterns were planned during the visit through pre-programmed path planning functions available in DJI Phantom 4. The first pattern (Pattern 1) was at lower elevations, designed to capture various viewing angles of the casemate. The second pattern (Pattern 2), flown at higher elevations, was designed to scan the entire fort. Both patterns were under the circular path. To ensure the success of the photogrammetry process, an overlap of approximately 70% to 80% was maintained between adjacent UAV images. Lastly, instead of letting the UAV camera directly face downward, we defined the camera angle (i.e., the angle between the normal direction of the image plane and the horizontal direction; denoted in Figure 4a) as about 70 degrees.

Figure 4a and b show the camera positions, marked as small blue patches, in both elevation and plan views. The casemate in Figure 4b cannot be seen directly from the figure because it is obstructed by blue patches. Both patterns of UAV image collection were achieved by a single flight performed by one UAV operator. As can be seen in the figures, a comprehensive scan of the casemate's surface can be captured.

### 4.2 Establishment of the 3D Point Cloud

We utilized Agisoft Metashape (Agisoft, 2020) for 3D dense point cloud reconstruction. To this end, the SfM-MVS algorithms were applied in the software to process the UAV images discussed in Section 4.1. The procedure began with detecting features, which are distinctive localized small image patches that contain unique pixel intensity distribution. Feature points consistently appear across multiple images, even if they are under different camera positions. Next, feature points from different images were paired together using feature-matching algorithms. These matched feature points served as correspondences to estimate the camera matrix and further reconstruct the sparse point cloud of the virtual scene, as shown by the blue dashed lines in Figure 4a. The creation of the sparse point cloud requires less computational cost. At this stage, a user can also effectively check the quality of the image alignment, pinpoint any incorrect alignments, and evaluate the distribution of feature points.

Based on the result of the sparse point cloud, the dense point cloud of Fort Soledad was then established (Figure 4c). This procedure involved adding more points between the initially matched feature points, substantially increasing the density and detail of the model. The established dense point cloud of Fort Soledad contained approximately 2.4 million 3D RGB points. Creating a dense point cloud is computationally demanding but provides a high-fidelity digital replica of the physical site. Lastly, we applied the segmentation techniques through CloudCompare (CloudCompare, 2024) to segment the casemate from the main point cloud. The point cloud of the casemate (Figure 4d) serves as the virtual entity of the historic structure in this study. For a detailed discussion of the SfM-MVS algorithms and 3D reconstruction technologies, especially in the context of cultural heritage, the reader is referred to the study performed by Kingsland (2020).

## 5. Validation Phase 2: Deterioration Monitoring

This section validates the feedback flow that links the virtual entity and the service (i.e., structural deterioration monitoring). Section 5.1 explains a two-stage point cloud alignment procedure to register two datasets of the casemate; Section 5.2 discusses the principle of computing cloud-to-cloud (C2C) distances and shows the results of C2C distances for a side wall of the casemate; Section 5.3 demonstrates the methodologies for simulating crack-like edges and true cracks and further illustrates the structural deterioration monitoring results against these two types of simulated features.

### 5.1 Point Cloud Alignment

As explained in Section 2.2, two point cloud datasets are required to monitor structural deterioration over time. To obtain the additional dataset, we revisited Fort Soledad to collect new UAV images, reconstruct the dense point cloud of the site, and truncate the casemate's point cloud, following the protocol explained in Section 4. Thereafter, both point clouds of the casemate were aligned together via a two-stage protocol described below:

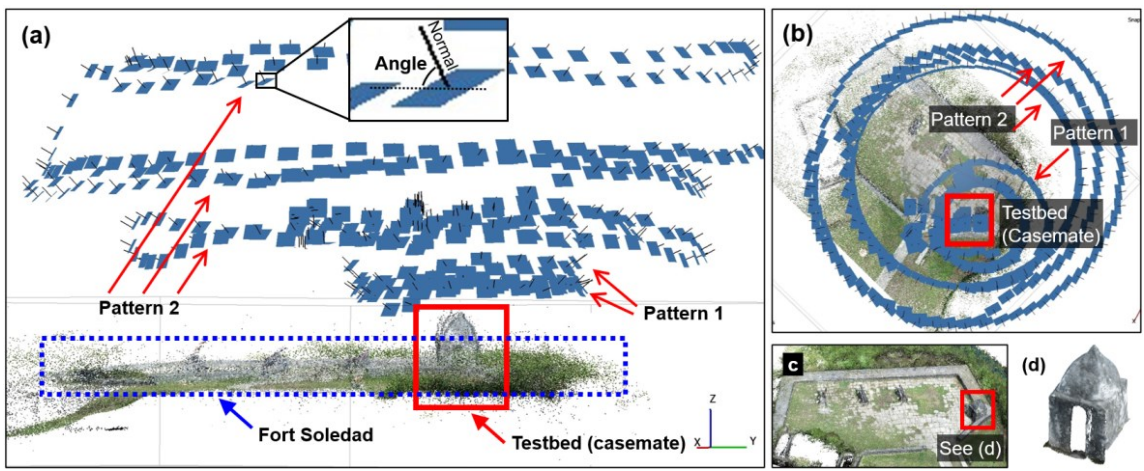

Figure 4. Virtual entity reconstruction: (a) and (b) show the camera positions (i.e., small blue patches) of UAV images from the side and plan views, respectively; (c) and (d) show the established 3D dense point cloud of Fort Soledad and the casemate, respectively.

First, point clouds were loaded into CloudCompare for rough alignment. As shown in Figure 5a, four correspondences (C1 to C4) were manually selected in each dataset for alignment. These correspondences served as the basis for initial rough alignment. Notice the selection of these correspondences is flexible, any other distinct features can be selected as correspondences if they can be identified by human eyes. Lastly, it is worth noting that registration errors may exist at this stage due to manual selections and they will be minimized in the second stage alignment.

Second, a fine adjustment was performed using the Iterative Closest Point (ICP) algorithm developed by Chen and Medioni (1992). The algorithm can refine the alignment by finding the closest point for each 3D point in the second visit's point cloud to a given point in the first visit's point cloud, minimizing point-to-point distance. Next, a geometric transformation matrix can be estimated using a root mean square error minimization, which can be formulated as:

$$E(R,t) = \min_{R,t} \sum_i \|p_i - (Rq_i + t)\|^2 \quad (1)$$

where $R$ and $t$ are the rotation and translation from the geometric transformation matrix; $E(R,t)$ is the error function; $p_i$ is a point from the point cloud in the first visit; and $q_i$ is a point from the point cloud from the second visit. Once established, this geometric transformation matrix was applied to the point cloud in the second visit to align it with the point cloud in the first visit.

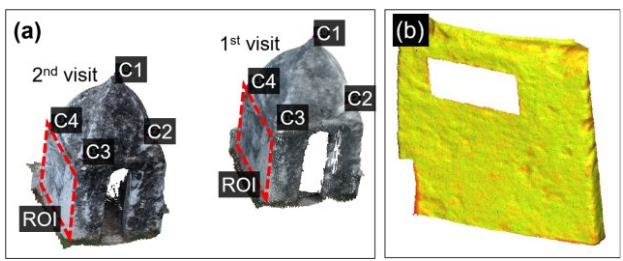

Figure 5. (a) Rough alignment of the casemate from two field visits; and (b) the C2C distances in the ROI.

### 5.2 C2C Distances

Once two point clouds from both visits are aligned through the procedure above, we calculated the cloud-to-cloud (C2C) distances between these two clouds to find time-varying changes caused by structural deteriorations. To this end, let us define a set $A = \{a_1, \ldots, a_p\}$ as the reference point cloud from the first visit, and a set $B = \{a_1, \ldots, a_q\}$, as the floating point cloud from the second visit. The Hausdorff distance (Huttenlocher, 1993) between these sets is formulated as:

$$H(A,B) = \max(h(A,B), h(B,A)) \quad (2)$$

where $h(A,B) = \max_{a \in A} \min_{b \in B} \|a - b\|$.

The function, $h(A,B)$, finds the largest single distance from any point in set $A$ to its closest point in set $B$, measuring the maximum possible discrepancy between the two datasets.

To better facilitate the C2C distance calculations, we defined a region of interest (ROI) on the side wall of the casemate, as depicted in the red dashed box in Figure 5a for both datasets. The purpose of selecting such an ROI is to quantify a localized area rather than the whole structure, such that any differential features caused by the deterioration can be easily identified.

Figure 5b shows the C2C distances in the ROI between two visits. Before performing C2C distance computation, we excluded any 3D points within the window area (i.e., the rectangular void in the figure) because these 3D points usually had low data quality caused by insufficient UAV image coverage.

### 5.3 Deterioration Monitoring

Because the two field visits were carried out under a short time interval, minimal structural deteriorations can be observed in Figure 5b. To evaluate our framework's ability to monitor time-varying changes that one would see in the real world, we developed methods for simulating both crack-like edges (i.e., fake cracks) and true cracks. Figure 6 and Figure 7 illustrate the methodologies used for creating these non-crack and crack features; while the validation of these methodologies was performed in CloudCompare which will be discussed later on in this section.

To simulate crack-like edges, we first define a narrow, slender-shaped element in the point cloud of the side wall, as shown in Figure 6a. This element is then segmented from the main wall, and colored black to enhance its visibility. Next, this colored slender element is merged back with the segmented wall, to form the final point cloud in the figure. Notice that this simulation involves no geometric alterations to the point cloud. Therefore, the crack-like edge is a slender element painted in black rather than a true crack.

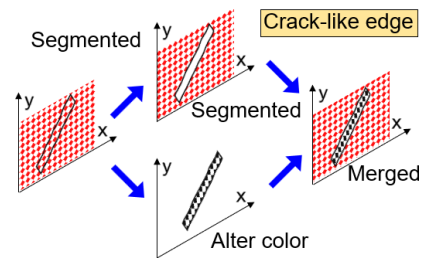

Figure 6. Methodologies to simulate crack-like edges.

The simulation of true cracks follows a similar procedure described above but with a critical modification: the geometric features of the slender element are changed. In particular, we slightly shift the element along its normal direction, which is perpendicular to the $xy$ plane, as shown in Figure 7. This element's movement along its normal direction is intended to simulate the penetrating behavior of cracking in stone walls, reflecting the depth that true cracks would exhibit. This adjustment ensures that the simulated cracks represent realistic depth, essential for accurate deterioration monitoring.

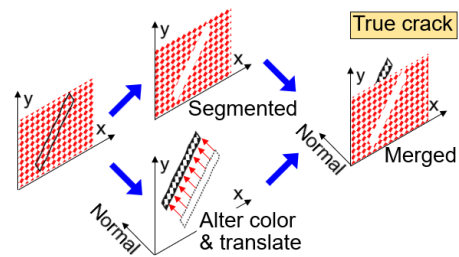

Figure 7. Methodologies to simulate true cracks.

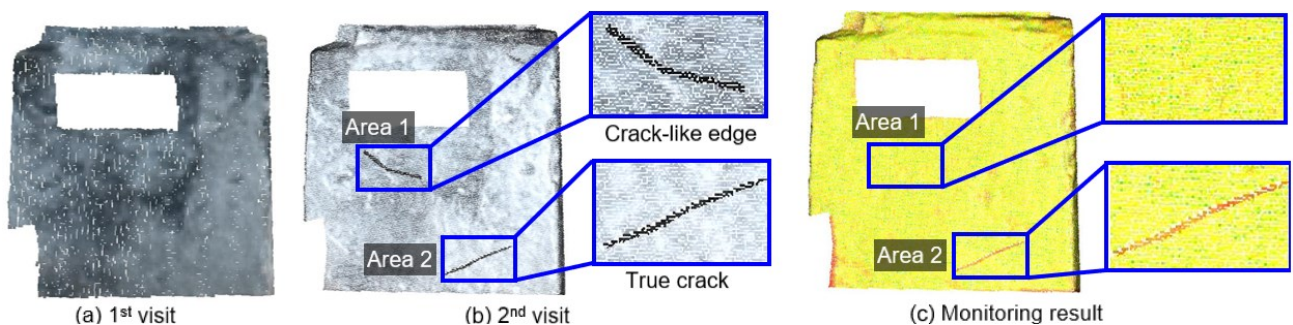

Figure 8. Crack monitoring result: (a) the point cloud from the first visit; (b) the point cloud from the second visit; and (c) C2C distances between the point clouds from two visits.

Figure 8 illustrates the validation procedure. First, we applied the crack simulation methods described earlier, introducing one simulated crack-like edge and one simulated true crack to the side wall of the casemate from the second visit. The results of these simulations are shown in Figure 8b. The C2C distances of two point clouds from the first visit (Figure 8a) and the second visit (Figure 8b) are calculated, showing in Figure 8c. As can be seen from Areas 1 and 2 in Figure 8b, the texture of the crack-like edge and true crack are very similar, making them difficult to distinguish even with human eyes. Nevertheless, our proposed method effectively identifies the true crack, as shown in Area 2 from Figure 8c, while correctly avoiding the detection of the false positive result as shown in Area 1 from Figure 8c.

## 6. Validation Phase 3: Decision Making

This section discusses the feedback flow that links the service (i.e., structural deterioration monitoring) and the physical entity. As can be seen from the results in Sections 4 and 5, our proposed method effectively detects damage in the stone wall of the casemate. In addition, our method is scalable and can be adapted to meet the needs of various stakeholders for their decision-making. For example, instead of finding the local scale cracks in this study, the method can be expanded to identify damage such as loosened stones and uneven settlement at mesoscale or large scales. Lastly, the monitoring approach can be applied to other parts of the casemate, such as the dome, or other historic structures. Readers are referred to Kong and Hucks, 2023 for a case study in monitoring the deterioration of a historic bridge.

This adaptability of our approach in different scales and types of structures would be crucial for decision-making. By providing detailed and scalable monitoring data on the condition of the historic structure, our method enables stakeholders to prioritize repairs, plan maintenance schedules, and allocate resources more effectively. For example, early detection of minor cracks can prompt timely interventions that prevent further damage, reducing long-term repair costs and enhancing structural integrity. Additionally, by understanding the broader implications of mesoscale and large-scale deteriorations, such as uneven settlement, stakeholders can make strategic decisions that address underlying issues.

## 7. Discussion

While there is no consensus on the exact definition of a digital twin in the literature, we embrace the ongoing debates and discussions about what constitutes a digital twin and how it can help us achieve our research goals. As discussed in Section 2.1, our conception of a digital twin contains five key components: physical entity, virtual entity, service, data, and connection. Although we highlighted the key distinctions between our digital twin framework and 4D modeling in Section 2.5, here we further differentiate our approach from other studies.

Firstly, we argue that the digital twin is more than just a digital model; it should be beyond simple digital modeling. Indeed, while the 3D dense point cloud model of the casemate in this study captures the geometric and color textures of the physical entity, the point cloud itself should not be considered a digital twin. To qualify as a digital twin, the system must incorporate the aforementioned five key components in the feedback loop, serving a specific service objective, which is the deterioration monitoring in this study.

Secondly, we argue that while real-time feedback is the preferred feature for a digital twin, not a requirement. Digital twin applications in manufacturing, such as those described by Thelen et al. (2022), typically emphasize real-time feedback from the virtual entity to the physical entity, often implemented through manufacturing plant control systems. However, given the context of heritage preservation, whether monitoring results can be synchronously looped back to influence the physical entity may not be essential and highly depends on the monitoring technologies employed. For instance, if a contact-based sensor network could be deployed, real-time monitoring and decision-making would be achievable. However, this could also result in increased implementation costs for property owners. Therefore, digital twin applications must be considered carefully within the specific constraints and needs of cultural heritage preservation.

## 8. Conclusion

In this study, we introduced an innovative digital twin framework for monitoring time-varying deteriorations of historic structures. We began by reviewing existing literature to identify the research gap, which is assessing time-varying changes in historic structures. Next, a five-component digital twin framework was developed. The proposed framework was then validated through a casemate at Fort Soledad on the island of Guam. We discussed the processes in validation including reconstructing the virtual entity of the casemate in Phase 1, monitoring structural deterioration in Phase 2, and the usage of monitoring results for decision-making in Phase 3.

The findings from this study confirm that our digital twin framework can effectively monitor the deterioration of historic structures, which is a critical need in the cultural heritage preservation community. As illustrated in the results of this paper, our work supports stakeholders in making sound decisions that enhance structural integrity. We hope that this research will contribute to the existing body of knowledge across the domains of photogrammetry, digital twin, and cultural heritage preservation; and prove valuable to our peers in these fields.

## 9. Future Work

Future work will focus on expanding the proposed digital twin framework from historic structure preservation to broad engineering applications such as civil infrastructure inspections. Findings in the field of computer vision-based structural health monitoring such as detecting steel cracks (Kong and Li, 2019) and loosened bolts (Kong and Li, 2018) will be investigated for their potential to be integrated into our framework.

## Funding

The study received partial support from the New Faculty Research Start-Up Fund from the Lyles College of Engineering at California State University, Fresno. However, any opinions, findings, conclusions, or recommendations expressed in this study are those of the author and do not necessarily reflect the views of the funder.

## Acknowledgments

The UAV and other equipment used in this study were partially funded by two seed grants from NASA Guam EPSCoR (# 80NSSC19M0044) and NSF Guam EPSCoR (#1457769). We thank Dr. Yushin Ahn at California State University, Fresno for the fruitful discussion on the digital twin; and anonymous reviewers for their constructive comments that improved the quality of this paper. Lastly, we acknowledge the assistance of ChatGPT in the proofreading process of this manuscript.